\documentclass{article}

\usepackage{microtype}
\usepackage{graphicx}
\usepackage{subcaption}
\usepackage{booktabs} % for professional tables
\usepackage{tikz}

\definecolor{pilotblue}{HTML}{4C78A8}
\definecolor{pilotgray}{HTML}{E2E5E9}
\definecolor{pilotink}{HTML}{4A4A4A}

\usepackage{hyperref}

\usepackage[accepted]{icml2026}

\usepackage{amsmath}
\usepackage{amssymb}
\usepackage{mathtools}
\usepackage{amsthm}

\usepackage[capitalize,noabbrev]{cleveref}

\usepackage{array}

\theoremstyle{plain}

\theoremstyle{definition}

\theoremstyle{remark}

\usepackage[disable,textsize=tiny]{todonotes}
\icmltitlerunning{Position: Legal Hallucination Is Warrant Failure}

\begin{document}

\twocolumn[
  \icmltitle{Position: Legal LLM Hallucination Should Be Evaluated as Failure of Legal Warrant}

  % It is OKAY to include author information, even for blind submissions: the
  % style file will automatically remove it for you unless you've provided
  % the [accepted] option to the icml2026 package.

  % List of affiliations: The first argument should be a (short) identifier you
  % will use later to specify author affiliations Academic affiliations
  % should list Department, University, City, Region, Country Industry
  % affiliations should list Company, City, Region, Country

  % You can specify symbols, otherwise they are numbered in order. Ideally, you
  % should not use this facility. Affiliations will be numbered in order of
  % appearance and this is the preferred way.
  \icmlsetsymbol{equal}{*}

  \begin{icmlauthorlist}
    \icmlauthor{Maksym Taranukhin}{ubc,vector}
    \icmlauthor{Vered Shwartz}{ubc,vector,cifar}
  \end{icmlauthorlist}

  \icmlaffiliation{ubc}{University of British Columbia, Vancouver, BC, Canada}
  \icmlaffiliation{vector}{Vector Institute, Toronto, ON, Canada}
  \icmlaffiliation{cifar}{CIFAR AI Chair}

  \icmlcorrespondingauthor{Maksym Taranukhin}{maksymt@cs.ubc.ca}
  \icmlcorrespondingauthor{Vered Shwartz}{vshwartz@cs.ubc.ca}

  % You may provide any keywords that you find helpful for describing your
  % paper; these are used to populate the "keywords" metadata in the PDF but
  % will not be shown in the document
  \icmlkeywords{legal AI, hallucination, retrieval-augmented generation, evaluation, access to justice}

  \vskip 0.3in
]

% this must go after the closing bracket ] following \twocolumn[ ...

% This command actually creates the footnote in the first column listing the
% affiliations and the copyright notice. The command takes one argument, which
% is text to display at the start of the footnote. The \icmlEqualContribution
% command is standard text for equal contribution. Remove it (just {}) if you
% do not need this facility.

% Use ONE of the following lines. DO NOT remove the command.
% If you have no special notice, KEEP empty braces:
\printAffiliationsAndNotice{}  % no special notice (required even if empty)
% Or, if applicable, use the standard equal contribution text:
% \printAffiliationsAndNotice{\icmlEqualContribution}

\begin{abstract}
In this position paper, we argue that legal LLMs' hallucinations should be evaluated as a failure of legal warrant rather than as factual inaccuracy or citation failure. We define claim-authority warrant as the context-sensitive relation between a consequential legal claim and authority that exists, applies to the relevant jurisdiction, is current for the date of analysis, has the legal status represented by the system, and supports the proposition asserted. Warranted legal generation is the broader system behavior that answers, narrows, asks, warns, corrects a false premise, or abstains according to that relation. The falsifiable prediction is that warrant metrics reveal material failures that answer accuracy, citation existence, generic attribution, LegalHalBench-style statute relevance, and CitaLaw-style sentence-citation alignment can miss. We sharpen this claim with a side-by-side comparison item and a small, reproducible pilot over public-rule tests. We then specify benchmark records, claim boundaries, support labels, mixed response-policy scoring, risk weights, annotation reliability reporting, and jurisdiction-specific authority ontologies. The result is a concrete research agenda for evaluating legal AI systems by whether their consequential claims are licensed by law.
\end{abstract}

\section{Introduction}
\label{sec:introduction}

Legal hallucination is often introduced as the invention of cases. That is the simplest failure, not the whole problem. In \textit{Mata v. Avianca}, lawyers filed non-existent cases and quotations generated by ChatGPT, leading to Rule 11 sanctions \citep{mata2023avianca}. More recent public incidents include filings or proposed orders with inaccurate citations, misstatements, fictitious or misattributed authority, and AI-assisted drafting failures \citep{reuters2026sullivan,reuters2026georgia}. These events mix fabrication with a subtler defect: a legal proposition can be attached to a real source and still not be licensed by that source.

The research evidence points in the same direction. General-purpose LLMs hallucinate on legal knowledge questions, vary across courts and time periods, accept false premises, and often lack calibrated awareness of their errors \citep{dahl2024large}. AI legal research tools that use retrieval reduce but do not eliminate incorrect or misgrounded answers \citep{magesh2025hallucination}. Legal retrieval benchmarks show that finding candidate authority remains difficult when answers depend on facts, exceptions, rule hierarchy, and source treatment \citep{zheng2025reasoning}.

Therefore, this position paper argues that \textbf{legal LLM hallucination should be evaluated as a failure of legal warrant.} A legal claim is not reliable because it sounds plausible or carries a real citation. A blog post, dissent, district court case, agency FAQ, and controlling statute do not warrant the same kind of claim. Legal reliability depends on whether authority licenses a proposition under the relevant jurisdiction, date, forum, procedural posture, source status, and support relation. This view builds on argumentation theory and legal reasoning about authority \citep{toulmin1958uses,schauer2009thinking}, but it changes ML evaluation. The object to evaluate is the relation between each consequential legal claim and the authority offered, retrieved, omitted, or shown to be unavailable.

\begin{figure*}[t]
\centering
\includegraphics[width=\textwidth]{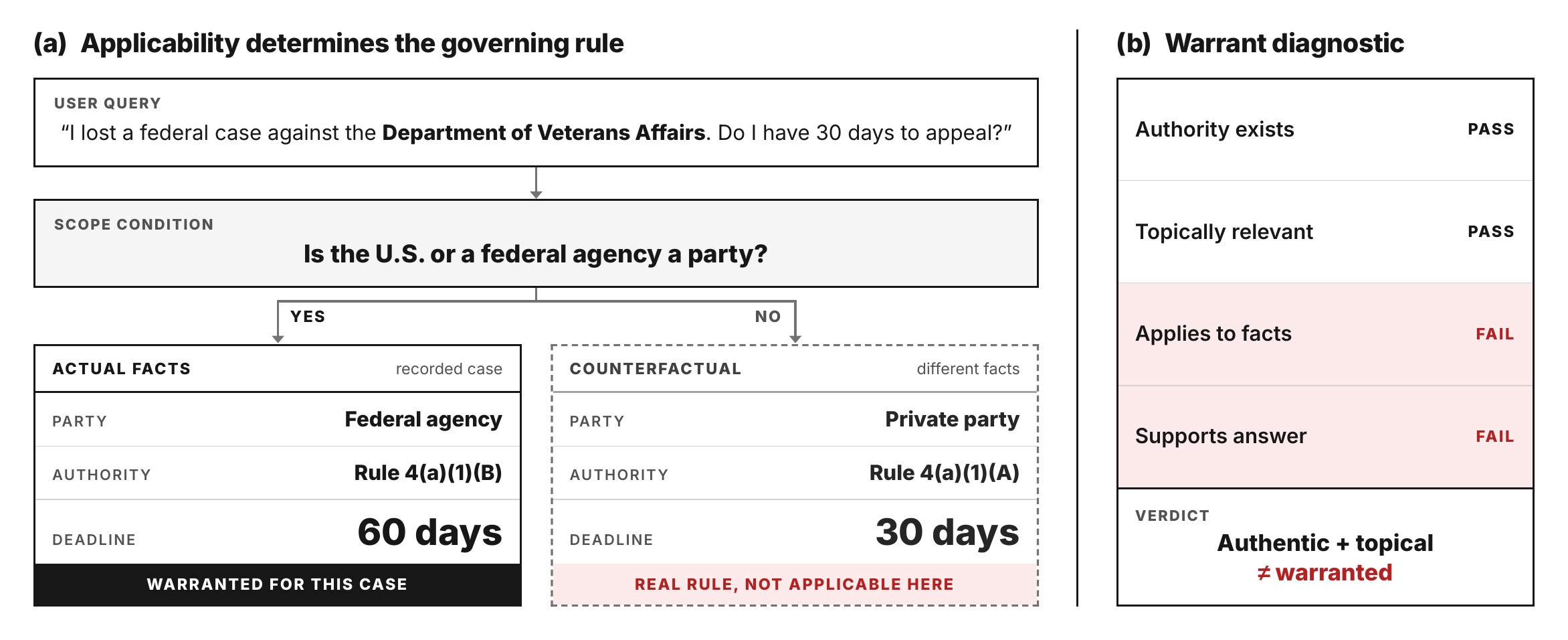}
\caption{Example of a legal warrant for a federal appeal deadline. (a) The recorded facts make Rule~4(a)(1)(B)'s 60-day period applicable while Rule~4(a)(1)(A)'s 30-day period governs only the private-party counterfactual. (b) Although the 30-day authority exists and is topically relevant, it does not apply to the recorded facts or support the answer.}
\label{fig:runningexample}
\end{figure*}

\Cref{fig:runningexample} shows an example that makes the target concrete. A user asks whether they have 30 days to appeal a federal civil judgment against the Department of Veterans Affairs. An answer citing Federal Rule of Appellate Procedure 4(a)(1)(A) for a 30-day rule uses a real, topical source, but it is overbroad. Rule 4(a)(1)(B) gives 60 days when the United States or a federal agency is a party. By contrast, a private-party version of the prompt would make Rule 4(a)(1)(A) the more relevant starting point \citep{frap2025}. 

This paper makes four contributions, which also define its structure. First, it introduces the \textbf{C}laim-\textbf{L}aw \textbf{A}uthority \textbf{W}arrant (CLAW) framework (\Cref{sec:warrant}), which defines legal warrant as an operational target and removes a circular definition by separating claim-authority warrant from response-policy adequacy. Second, it maps recent legal benchmarks such as LegalHalBench and CitaLaw onto the warrant target, showing that they measure necessary components of warrant but do not by themselves establish it (\Cref{sec:warrant,sec:limitations}) \citep{hu2025legalhalbench,zhang2025citalaw}. Third, it gives a proof-of-concept annotation pilot that shows how citation existence and topicality can pass while the warrant fails (\Cref{sec:pilot}). Fourth, it supplies a concrete roadmap for warrant benchmarks and warrant-aware systems, including a minimum viable warrant suite (\Cref{sec:benchmark,sec:a2j,sec:minimum}).

\section{The CLAW Framework}
\label{sec:warrant}

Evaluating warrant requires a precise statement of what is being checked, for which unit of text, and in which legal context. The CLAW framework uses \emph{claim-authority warrant} for a ternary relation $W(c,a,k)$ among a claim $c$, an authority or marked absence of authority $a$, and legal context $k$. The context records jurisdiction, forum, date of analysis, procedural posture, user type, source corpus, and source-status ontology. A claim-authority pair is warranted when the source exists, the source supports the proposition, the source applies in the recorded legal context, the source is current for the date of analysis, and the source has the status represented by the system. \emph{Response-policy adequacy} is separate. It asks whether the system should answer, narrow, ask for missing facts, warn, correct a false premise, or abstain. \emph{Warranted legal generation} is the composite system goal: make only warranted consequential claims and choose a response policy suited to missing, weak, or contested warrants.

The unit is a \emph{consequential legal claim}. We propose a master boundary rule (see details in \Cref{app:claim_boundary_rules}). A generated statement is consequential if changing its truth value would plausibly change a user's legal action, risk assessment, deadline, remedy, argument, right, duty, burden, forum choice, or need to seek professional help. For public-user tools, the test is whether a reasonable lay user might act differently because of the statement. For lawyer-facing tools, the test is whether the statement would need authority in a memorandum, brief, advice letter, or research note. Background definitions are consequential only when they are used to support advice, triage, or a legal conclusion. Domain guides should instantiate this master rule with examples and counterexamples.

\begin{table*}[t]
\caption{Legal-warrant failure modes and reasons common evaluation scores may miss them; a single answer may exhibit multiple modes.}
\label{tab:failures}
\centering
\small
\setlength{\tabcolsep}{4pt}
\begin{tabular}{p{0.16\textwidth}p{0.42\textwidth}p{0.35\textwidth}}
\toprule
\textbf{Failure mode} & \textbf{What fails} & \textbf{Why common scores miss it} \\
\midrule
\textit{Existence} & A case, statute, rule, quotation, docket, or citation does not exist. & Citation checks catch this, but often treat it as the whole problem. \\
\textit{Attribution} & A real source is cited for a proposition it does not support. & The answer appears grounded because the source is genuine and topical. \\
\textit{Authority \newline status} & Persuasive authority, dicta, a dissent, a trial ruling, or secondary material is represented as controlling law. & Generic support evaluation does not consider the precedential value or legal authority of the source. \\
\textit{Jurisdiction} & The answer imports law from the wrong forum, agency, court hierarchy, or legal system. & The proposition may be true somewhere else. \\
\textit{Temporality} & The rule is obsolete, amended, overruled, stayed, or not yet effective. & Static labels hide the date for which a rule is valid. \\
\textit{Procedural posture} & The answer ignores stage, burden, standard of review, remedy, or filing posture. & The statement may be abstractly true but unusable. \\
\textit{False-premise compliance} & The system accepts an incorrect legal assumption rather than correcting it. & The answer can be coherent relative to the prompt while failing the legal need. \\
\textit{Epistemic overreach} & The system gives confident advice when authority is missing, conflicted, or facts are unknown. & The answer may be partly true, but lacks warranted scope. \\
\bottomrule
\end{tabular}
\end{table*}

\Cref{tab:failures} separates failures often collapsed into one hallucination rate, in contrast to coarser legal-AI risk taxonomies \citep{buchicchio2024design}. Its eight rows are diagnostic categories for stress-test design and reporting, not eight independent labels assigned to every span. Annotators instead record a compact schema for each claim-authority-context tuple: source existence and status, jurisdiction-time-posture metadata, one support label, a response-policy vector, and a risk weight. The categories are derived from those fields, which keeps annotation modular and permits later consolidation when dimensions are redundant. A system can have perfect citation existence and still fail attribution, source status, time, jurisdiction, or posture. It can also avoid false claims by refusing everything while withholding warranted information. Warrant evaluation therefore needs both negative labels for unsupported claims and positive labels for useful narrowing. \Cref{tab:coverage} compares this record with adjacent legal benchmarks. Epistemic overreach from \Cref{tab:failures} is operationalized below under response policy, since confident advice without supporting authority is a policy failure.

\begin{table*}[t]
\caption{Coverage of warrant dimensions in LegalHalBench, CitaLaw, and the proposed warrant record; ``partial'' denotes coverage of some instances without a required field for every consequential claim.}
\label{tab:coverage}
\centering
\small
\setlength{\tabcolsep}{4pt}
\begin{tabular}{p{0.18\textwidth}p{0.23\textwidth}p{0.23\textwidth}p{0.27\textwidth}}
\toprule
\textbf{Dimension} & \textbf{LegalHalBench} & \textbf{CitaLaw} & \textbf{Warrant record} \\
\midrule
\textit{Source existence} & Strong for named statutes through non-hallucinated statute rate. & Present through retrieved law articles and cases. & Required for every cited or offered authority, including quotations and pinpoint text. \\
\textit{Topical relevance} & Strong through statute relevance. & Strong through retrieval and citation attachment. & Necessary but not sufficient. A topical source may still fail to support. \\
\textit{Claim support} & Scored as legal claim truthfulness. & Scored through sentence-citation entailment and syllogism components. & Labeled per consequential claim as direct, inferential, partial, contradiction, no address, out of scope, or unsettled. \\
\textit{Jurisdiction and time} & Partial when encoded by dataset scope. & Partial when encoded by the corpus and question. & Explicit fields in each record, with old and new law preserved as temporal tests. \\
\textit{Authority status} & Limited for statute-centered items. & Partial through the law article versus precedent distinction. & Jurisdiction-specific source ontology with binding force, institutional rank, treatment, and role in reasoning. \\
\textit{Procedural posture} & Partial through scenario text. & Partial through the circumstances component. & Required context field when posture changes, remedy, burden, deadline, or standard. \\
\textit{Response policy} & Helpfulness or style can reward usable answers. & Style and correctness are measured. & Multi-label behavior score for answer, narrow, ask, warn, abstain, and false-premise correction. \\
\bottomrule
\end{tabular}
\end{table*}

We predict that warrant metrics will produce the largest gaps over adjacent metrics for attribution, authority status, procedural posture, and response policy. A source can be real, relevant, and sentence-aligned while supporting only a narrower proposition; existing metrics often do not test whether a source is binding, persuasive, secondary, dicta, dissent, or negatively treated; and relevance or citation alignment can reward an answer that should instead ask, narrow, warn, correct, or abstain. Procedural-posture gaps should be especially large for deadlines, remedies, burdens, standards of review, and litigation stage. We expect medium-to-high gaps for jurisdiction and time, which dataset scope can conceal even though deployed systems answer across borders and dates, and smaller gaps for existence and topicality, which current legal citation benchmarks already target well.

These predictions can be audited on existing outputs by adding metadata and support labels. Attribution audits should relabel cited sentences as direct, inferential, partial, contradiction, no address, out of scope, or unsettled; authority-status audits should add source-type and treatment metadata and compare the force claimed with the source's local legal force. Jurisdiction-and-time audits should swap jurisdiction, effective date, or forum while keeping the topical source visible, and posture audits should record whether the output conditions the rule on posture and whether the cited authority applies to that posture. Response-policy audits should label answer, narrow, ask, warn, abstain, and correct acts as a policy vector and apply vetoes to unsafe unsupported claims. Existing statute-existence, relevance, and citation-validity checks remain shared submetrics.

The running example in \Cref{fig:runningexample} appears successful under adjacent scoring objects even though its consequential advice remains unsupported. Citation existence passes because FRAP 4(a)(1)(A) exists, but existence alone cannot determine whether another provision controls the facts. LegalHalBench-style relevance is high because the rule concerns civil appeal deadlines, but topical relevance does not establish that the rule supports this user's deadline. CitaLaw-style sentence-citation alignment passes or partially passes because the narrower statement ``Rule 4(a)(1)(A) says 30 days'' is entailed, but sentence-level alignment can still pass when the final advice is broader than the cited proposition.

The warrant record instead fails both support and response policy because it asks which authority controls for the recorded party type and date. Here, Rule 4(a)(1)(B) governs when a federal agency is a party, so a warranted response should narrow the claim, cite the 60-day provision, and avoid an unqualified ``yes.'' This divergence motivates the pilot below, which tests whether the separation persists across a broader set of stress tests.

\section{A Pilot Annotation}
\label{sec:pilot}

To make the empirical claim concrete, we constructed and labeled a reproducible test. The six prompts are manually designed stress tests. Each prompt targets one or more failure modes in \Cref{tab:failures}: near-miss authority (a real and topical source that governs a slightly different situation than the user's), false-premise compliance, temporal treatment, procedural posture, jurisdictional underspecification, and source-status overreach. The jurisdictional items reflect evidence that hallucination rates vary across places and jurisdictions for place-based legal queries \citep{curran2025place}. One bankruptcy item is inspired by published RAG audit examples where legal research tools overstated jurisdictionality from topical bankruptcy material \citep{magesh2025hallucination}.

\begin{figure}[t]
\centering
\begin{tikzpicture}[font=\scriptsize, x=1cm, y=1cm]
  \newcommand{\pilotadjbar}[4]{%
    \node[anchor=east, text=pilotink] at (2.35,#1) {#2};
    \filldraw[fill=pilotgray, draw=pilotink, line width=0.35pt]
      (2.55,{#1-0.16}) rectangle ({2.55+0.0515*#3},{#1+0.16});
    \node[anchor=east, text=pilotink, font=\tiny]
      at ({2.55+0.0515*#3-0.08},#1) {#4};
  }
  \newcommand{\pilotwarrantbar}[4]{%
    \node[anchor=east, text=pilotink] at (2.35,#1) {#2};
    \filldraw[fill=pilotblue, draw=pilotink, line width=0.35pt]
      (2.55,{#1-0.16}) rectangle ({2.55+0.0515*#3},{#1+0.16});
    \node[anchor=west, text=pilotink, font=\tiny]
      at ({2.55+0.0515*#3+0.08},#1) {#4};
  }

  % Legend: color and fill both distinguish adjacent checks from warrant outcomes.
  \filldraw[fill=pilotgray, draw=pilotink, line width=0.35pt]
    (2.55,5.17) rectangle (2.83,5.35);
  \node[anchor=west, text=pilotink, font=\tiny] at (2.92,5.26) {Adjacent/necessary};
  \filldraw[fill=pilotblue, draw=pilotink, line width=0.35pt]
    (5.20,5.17) rectangle (5.48,5.35);
  \node[anchor=west, text=pilotink, font=\tiny] at (5.57,5.26) {Warrant-specific};

  % Shared 0--100 percent scale.
  \foreach \x in {0,50,100} {
    \draw[pilotink!25, line width=0.3pt]
      ({2.55+0.0515*\x},1.18) -- ({2.55+0.0515*\x},4.68);
    \node[anchor=north, text=pilotink, font=\tiny]
      at ({2.55+0.0515*\x},1.08) {\x};
  }

  \node[anchor=west, font=\scriptsize\bfseries, text=pilotink]
    at (0,4.72) {Output-level measures ($n=18$)};
  \pilotadjbar{4.18}{Existence}{100}{100\% (18/18)}
  \pilotadjbar{3.68}{Topicality}{100}{100\% (18/18)}
  \pilotwarrantbar{3.18}{Policy adequacy}{38.9}{38.9\% (7/18)}

  \node[anchor=west, font=\scriptsize\bfseries, text=pilotink]
    at (0,2.63) {Claim--authority pairs ($n=42$)};
  \pilotadjbar{2.09}{Sentence support}{66.7}{66.7\% (28/42)}
  \pilotwarrantbar{1.59}{Full warrant}{47.6}{47.6\% (20/42)}

  \node[anchor=north, text=pilotink] at (5.125,0.62) {Pass rate (\%)};
\end{tikzpicture}
\caption{Pilot pass rates by scoring unit, with exact counts; the adversarial sample evaluates label separability rather than prevalence, and output-level and claim--authority-pair measures use different denominators.}
\label{fig:pilotresults}
\end{figure}
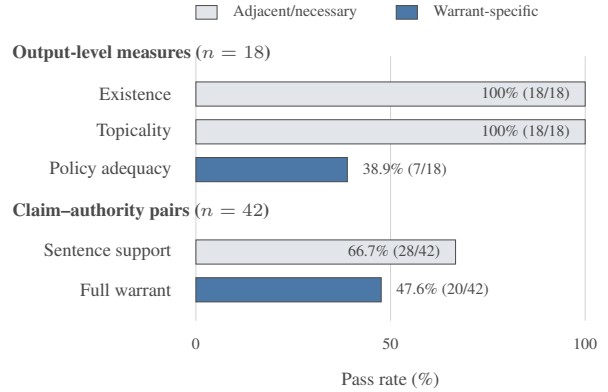

For each prompt, we manually wrote three short candidate outputs rather than generating them with a model. The output patterns are a default-rule answer with a real topical source, a broad refusal or generic disclaimer, and a warranted-narrowing answer. A warranted-narrowing answer gives only the part licensed by the available authority, states the condition under which it applies, asks for missing legally operative facts when needed, and warns against treating an unverified condition as settled. This produces 18 outputs. The 42 claim-authority pairs come from extracting every consequential claim in those outputs and linking each claim to the source it offers or implicitly relies on. \Cref{app:pilotitems} gives the prompts, output templates, outputs, and labels. The pilot does not evaluate any deployed model; it shows how candidate answers would be scored. It tests whether the proposed labels are operational and whether they separate warrant from adjacent metrics; \Cref{fig:pilotresults} reports the pass rates by measure and scoring unit.

Three observations follow. First, in this adversarial pilot, citation existence and topicality are deliberately easy to satisfy because every default answer cites a real, topic-adjacent source. The measured gap, therefore, isolates the harder dimensions: attribution, temporal or procedural fit, authority status, and response policy. Second, partial support is common. A source may support ``ordinary civil appeals are due in 30 days'' while not supporting ``your appeal is due in 30 days.'' Third, broad refusal is not a solution. It avoids unsupported claims, but it fails to give warranted procedural information. The private-party near miss also shows that a default-rule answer can be adequate when the facts actually satisfy the default. This pilot is deliberately small, intended to make the falsifiable agenda testable. A future release can replace the hand-written outputs with outputs from real systems and ask whether the gap persists.

\section{A Roadmap for Warrant Benchmarks}
\label{sec:benchmark}

In this section, we turn the warrant target into a benchmark-building protocol and assess the feasibility and cost of that protocol. The goal is to evaluate generated legal answers by decomposing them into consequential claims, authorities, context metadata, support relations, response-policy acts, and user-risk weights. The same records can also support auxiliary modeling tasks, such as claim extraction or support-label prediction, but the primary benchmark use is open-system evaluation, i.e., a system produces an answer, annotators or validated tools build a warrant record for that answer, and scores are reported by dimension.

\textbf{Grading pipeline.} Evaluating an open system therefore follows an explicit workflow. The system under test produces an answer to the benchmark prompt. Consequential claims are then extracted from the answer, manually or with model assistance, and audited by a legally trained reviewer against the claim-boundary rule. Each audited claim is linked to the authority it offers or implicitly relies on, and the link is resolved against the frozen source snapshot for the item's analysis date. Finally, annotators assign the support and response-policy labels defined below. Model assistance can extend beyond extraction: an LLM-as-judge can propose support labels at low cost, but published audits of legal AI tools show that automated judges share the failure modes under evaluation, so judge proposals for high-risk claims require expert confirmation \citep[\S 5.3]{magesh2025hallucination}. The pipeline's expected errors are also predictable. False positives arise when a judge or annotator accepts topical but non-supporting authority or overlooks a defeating condition, and false negatives arise when a strict reading rejects legitimate inferential support or when the snapshot omits an authority the system validly relied on. Reporting audit rates for extraction and linking alongside label agreement makes these error sources visible.

A benchmark item should contain more than a prompt and an answer label. It should record the user scenario, jurisdiction, forum, date of analysis, procedural posture, user type, source corpus, gold issue, acceptable authorities, source status, support labels, response-policy labels, and risk weights. The scored object is a tuple $(q,c,a,m,r)$, where $q$ is the task context, $c$ is a consequential claim, $a$ is an authority or marked absence of authority, $m$ is metadata about jurisdiction, date, forum, posture, and source status, and $r$ is the response policy. The same surface answer can receive different labels when jurisdiction or date changes.

\textbf{Support labels.} Each consequential claim should be linked to an authority or to a label for which no legal authority is needed. Direct support means the source states the proposition, and its conditions are met. Inferential support means the source licenses the proposition through a stated legal inference, such as a definition, exception, or incorporated rule. Partial support means the source supports a narrower proposition or only one required condition. Contradiction means the source refutes the claim. No address means the source is topical but silent on the proposition. Out of scope means the source is legally inapplicable because of jurisdiction, time, status, or posture. Unsettled means reasonable authority conflicts or no controlling authority resolves the issue. \Cref{tab:supportlabels} gives the decision rule and a typical legal example for each label. Existence itself admits strictness levels: a cited source may exist as a document while the pinpoint provision, quotation, or role it is cited for does not. Benchmark guides must state which strictness level their existence check uses; calibrating that choice is future work.

\begin{table*}[t]
\caption{Decision rules and legal examples for support labels applied to claim--authority--context tuples.}
\label{tab:supportlabels}
\centering
\small
\setlength{\tabcolsep}{4pt}
\begin{tabular}{p{0.16\textwidth}p{0.42\textwidth}p{0.35\textwidth}}
\toprule
\textbf{Label} & \textbf{Decision rule} & \textbf{Typical legal example} \\
\midrule
\textit{Direct support} & The source states the proposition, and all stated conditions are satisfied by the scenario. & A rule states the exact filing period for the recorded forum and party type. \\
\textit{Inferential support} & The source licenses the proposition through an explicit legal inference using definitions, cross-references, or exceptions. & A statute defines a covered person and another section imposes the duty. \\
\textit{Partial support} & The source supports a narrower proposition or only one element of the generated claim. & A default rule supports a deadline for ordinary civil cases, but not the federal-agency exception. \\
\textit{Contradiction} & The source refutes the proposition or supplies a rule that makes it false in context. & Current law rejects a standard that the answer says controls. \\
\textit{No address} & The source discusses the topic, but does not speak to the proposition. & An agency FAQ describes appeals generally but does not state the user's deadline. \\
\textit{Out of scope} & The source is legally inapplicable because of jurisdiction, date, forum, posture, or source status. & A California trial order is offered as controlling law in a British Columbia housing matter. \\
\textit{Unsettled} & Valid sources conflict or no controlling source resolves the claim. & Split authority on a novel statutory interpretation. \\
\bottomrule
\end{tabular}
\end{table*}

A good-faith legal argument can be warranted even when it is not legally correct in the sense of winning. The output must accurately state current law, mark the requested extension or change as an argument, and disclose contrary authority. The failure is not losing the argument. The failure is representing an analogy, dissent, policy preference, or proposed extension as controlling law.

\textbf{Mixed response policies.} The policy label should be a multi-label vector over answer, narrow, ask, warn, abstain, and correct, with definitions given in \Cref{tab:policies}. A response can properly combine acts, such as answering a general process question, narrowing the deadline claim, asking for jurisdiction \citep{taranukhin2026infogatherer}, and warning that local rules may matter. Benchmark reports should compute micro and macro $F_1$ between the output's policy vector and the gold policy vector. They should then apply deterministic veto rules for unsafe behavior. For example, an output receives zero policy adequacy for a high-risk item if it states an unsupported deadline, eligibility rule, forum choice, waiver consequence, custody consequence, detention consequence, or immigration consequence as settled law. It receives capped credit when it asks a useful question but also overclaims, or when it refuses everything although a narrower warranted answer is available. These vetoes are pre-registered in the annotation guide, so they are part of the metric rather than post-hoc judgment.

\textbf{Risk weights.} User-risk weighted warrant error should use a pre-registered harm rubric rather than ad hoc weights. Low risk covers background statements unlikely to change the action. Medium risk covers claims that may affect planning or document preparation. High risk covers deadlines, forum choice, eligibility, waiver, custody, housing loss, detention, immigration status, criminal exposure, or irreversible procedural steps. Weights should be assigned by task designers and legally trained reviewers, reported by domain, and accompanied by sensitivity analysis. Validation against observed user harm is a later empirical task, and is not a prerequisite for useful benchmark construction.

\textbf{Reliability.} Annotation should be staged: record context, extract claims, link candidate authority, label support, then label response policy. Benchmarks should publish agreement by dimension rather than one aggregate number. Existence and pinpoint matching may have high agreement. Inferential support, source treatment, and posture may not. That pattern is informative because it tells developers which parts of the legal warrant need better metadata or clearer annotation guides. LegalBench and LexGLUE show that legal tasks can be collaboratively annotated and benchmarked, but warrant annotation should preserve disagreement for unsettled law rather than force false consensus \citep{guha2023legalbench,chalkidis2022lexglue}.

\textbf{Warrant cards.} Evaluation should produce a warrant card for each output, with the minimum fields listed in \Cref{tab:warrantcard}. A lawyer-facing interface may expose the full card. A public-facing interface may translate it into plain language. The card is not a new disclaimer. It is a compact representation of what the system actually knows, what it assumes, and which claims remain unsupported.

\subsection{Feasibility, Cost, and Generalization}
\label{sec:feasibility}

\textbf{Scope.} A first warrant benchmark should be narrow. Good early domains include federal appellate deadlines, state housing repairs, immigration form triage, benefit eligibility thresholds, and administrative appeal windows. These tasks are compact enough for expert review and consequential enough to reveal failures. A realistic first release might contain 250 prompts, three seed outputs per prompt, and 1,500 to 3,000 claim-authority pairs. This scale is diagnostic, not exhaustive of the long tail of legal phrasing, domains, jurisdictions, or user circumstances. Coverage should be reported by stress-test family and expanded with paraphrase, counterfactual near-miss, and cross-jurisdiction variants, while claim-extraction precision and recall are measured on fresh system outputs. Seed outputs provide reproducible reference material for comparing metrics, training auxiliary extractors, and stress-testing annotation. New systems are evaluated by applying the same extraction, linking, and support-label protocol to their own outputs.

\textbf{Cost.} A first-pass annotator can draft claim spans and source links in 10 to 20 minutes per prompt when the domain and source corpus are narrow. Legal review and adjudication may add 20 to 40 minutes for difficult items. A 250-prompt release therefore has a rough budget of 125 to 250 expert hours, plus setup time for source snapshots and annotation guides. These estimates come from a narrow pilot, and genuinely hard items may exceed them. This is comparable to recent legal benchmark construction efforts that already report substantial lawyer time, such as LegalHalBench's more than 200 hours of professional review \citep{hu2025legalhalbench}. Costs can be reduced by model-assisted claim extraction, reusable authority graphs, and active sampling, but the final support and policy labels for high-risk claims should remain legally reviewed.

\begin{table}[t]
\caption{Minimum fields in a warrant card for a generated legal answer.}
\label{tab:warrantcard}
\centering
\small
\begin{tabular}{p{0.30\columnwidth}p{0.58\columnwidth}}
\toprule
\textbf{Field} & \textbf{Purpose} \\
\midrule
\textit{Legal context} & Jurisdiction, date, forum, domain, user type, and posture. \\
\textit{Claim ledger} & Consequential claims extracted from the answer. \\
\textit{Authority links} & Sources offered for each claim, with pinpoint locations. \\
\textit{Support label} & Direct, inferential, partial, contradicted, not addressed, out of scope, or unsettled. \\
\textit{Status metadata} & Binding force, source type, treatment, and effective date. \\
\textit{Response policy} & Answer, narrow, ask, warn, abstain, or correct. \\
\textit{Risk note} & Why an unsupported claim could harm the user. \\
\bottomrule
\end{tabular}
\end{table}

\textbf{Generalization.} The authority-status ontology must be jurisdiction-specific. The common-law labels in many U.S. examples, such as controlling, persuasive, dicta, dissenting, overruled, and superseded, are one instantiation. Civil-law systems may use code articles, regulations, constitutional decisions, administrative interpretations, jurisprudence constante, doctrinal commentary, and court decisions with different formal force. The general schema should therefore encode functional dimensions: source type, institutional rank, binding force, temporal effect, treatment status, and role in the reasoning. Local benchmark guides should map those dimensions to the legal system being tested. Cross-jurisdiction evaluation should compare whether systems respect the local ontology and not whether every system fits U.S. common-law categories.

\textbf{Source snapshots.} Legal materials change. A benchmark should store or identify the version of each source available at the analysis date. When the law changes, the old item can remain valid as a temporal test, and a new item can be added for the new date. Papers should state what metadata the system had access to. A system with proprietary citator treatment history should not be compared naively with one restricted to public text.

\textbf{Reporting.} Benchmark builders should resist a single leaderboard culture. Warrant data is most valuable when it reveals where systems break. A model might be strong on existence and weak on procedural posture. Another might ask good clarifying questions but omit contrary authority. Reporting these dimensions separately will make progress slower to summarize, but more useful for deployment. It will also reduce incentives to optimize for citation density or polished disclaimers at the expense of actual support.

\section{A Roadmap for Warrant-Aware Systems}
\label{sec:a2j}

Benchmarks are only half of the agenda. This section turns the warrant target into design guidance for systems, with access to justice as the motivating deployment setting, and closes with what should count as progress.

The access-to-justice setting is where warrant matters most. The Legal Services Corporation's 2022 Justice Gap Study reports that low-income Americans did not receive any or enough legal help for 92\% of their civil legal problems \citep{lsc2022justicegap}. AI systems may help explain processes, translate legal jargon, summarize documents, and triage issues. The risk is asymmetric. Lawyers can verify citations. Self-represented users may treat an answer as the best available legal guidance.

The target should be calibrated assistance instead of maximal refusal. A public-facing system should answer the parts it can warrant, state assumptions, ask for missing jurisdiction or facts, and refuse only the specific unsupported claim. For example, rather than saying only ``I cannot provide legal advice,'' it can say, ``I can explain the general filing steps, but I cannot state the deadline until I know the jurisdiction and the event that started the clock.'' Benchmarks should score this mixed behavior directly.

Access-to-justice tools need different benchmarks from lawyer-facing tools. Legal research assistants may assume the user is familiar with Shepardizing, checking citators, and distinguishing holdings from dicta. A public self-help tool cannot. Its evaluation should test whether the system prevents foreseeable user mistakes, such as filing in the wrong forum, missing a deadline, misdescribing a remedy, treating general information as personal advice, or relying on law from another jurisdiction. Multilingual systems should be judged by whether translated claims remain warranted in the local legal system, even when the English U.S. answer sounds plausible.

Early warrant suites should include false-premise, near-miss authority, temporal-shift, hierarchy-conflict, underspecification, source-omission, and cross-jurisdiction-transfer tests. \Cref{tab:stress} gives the full stress-test list.

Evaluation also changes system design. First, systems should maintain explicit claim ledgers before final generation. The generator should identify consequential claims, attach source candidates, verify support, and remove or narrow unsupported statements. Second, retrieval should be jurisdiction-time aware and should index source type, effective date, court hierarchy, agency, posture, and treatment history. Third, reranking should be authority-aware, including treatment-graph-aware reranking where treatment data exists. Fourth, support verification should check the proposition, the scope of the source, the status of the reasoning, and whether the answer overstates a rule. Fifth, interfaces should expose warrant in a role-appropriate form. A lawyer may want a table of claims, pinpoint sources, and treatment status. A public user may need a plain-language statement of what is assumed, what is known, and what requires local confirmation.

The warrant view turns legal hallucination into several ML problems. Consequential-claim extraction is not the same as atomic fact decomposition \citep{min2023factscore} because legal outputs contain background, caveats, source descriptions, and practical instructions. Legal support verification is not only NLI \citep{dagan2013recognizing} because rules interact with definitions, exceptions, burdens, standards of review, and facts from the prompt; treating it as generic entailment presupposes exactly the specialized legal knowledge whose verification is at issue. Authority representation requires source graphs that encode jurisdiction, hierarchy, enactment and effective dates, amendments, negative treatment, posture, and source type. Selective prediction should estimate confidence over support relations rather than surface fluency. These are tractable research problems, but they require benchmarks that expose the structure of legal justification.

The scope of the position is limited. Not every legal interaction requires a full warrant card. Translation, vocabulary explanation, and document summarization may need lighter records than deadline advice or litigation strategy. The position also does not require a model to resolve every hard legal question. Some questions are unsettled, fact-dependent, or governed by local practice outside the source corpus. In those cases, the correct behavior is to surface uncertainty and the verification path.

\subsection{What Counts as Progress}
\label{sec:progress}

Progress should be measured against baselines that separate fluency from warrant. Under warrant-based evaluation, a system improves when it produces fewer unsupported consequential claims, even when broad preference evaluators favor another system's prose or formatting. This distinction matters because human and model preference signals can favor surface formats and verbosity over responses of equal or better content quality \citep{zhang2024formatbias}. A system also improves if its citations are proposition-supporting, if it selects a scoped answer, question, warning, or uncertainty statement when appropriate, and if performance holds across jurisdictions, source types, legal domains, and user scenarios.

Overall accuracy can hide the most important failures. A model might perform well on federal appellate deadlines and poorly on state housing law. It might cite statutes accurately but misstate administrative deadlines. It might answer lawyer-facing research questions well while failing public-user triage. Warrant reports should therefore be disaggregated by domain, jurisdiction, source type, user type, date, risk tier, and stress-test family.

The warrant view also clarifies negative results. If a retriever finds the right source but the generator overstates it, the bottleneck is support verification. If the generator abstains whenever the law is local, the bottleneck is the response policy. If the model confuses binding authority with persuasive material, the bottleneck is source-status representation. If performance collapses after an amendment, the bottleneck is temporal indexing. This diagnostic framing is more useful for deployment than a single leaderboard score.

\section{Call to Action: A Minimum Viable Warrant Suite}
\label{sec:minimum}

The roadmap can begin with a deliberately small open release in one or two domains, following the scope and cost envelope of \Cref{sec:benchmark}. Benchmark builders should pair each prompt with a near-miss variant that changes jurisdiction, date, party type, posture, or user facts, then publish the source snapshot, annotation guide, risk rubric, and stress-test templates. Each item should include both a conventional answer key and a warrant card with claim spans, authority links, context fields, support labels, policy labels, risk tier, and annotator notes. This dual format permits direct comparison with existing accuracy, citation, and attribution metrics.

System developers should contribute outputs from at least one general LLM, one retrieval-augmented legal QA system, and one open model with a public prompt. Benchmark builders should add adversarial templates and human-written warranted references as audit targets, not prevalence estimates or perfect advice. Legal annotators should double-label a representative sample and report agreement by dimension, adjudicated labels, unresolved conflicts, and extraction and linking audit rates. The decisive test is whether warrant records change rankings or diagnoses for consequential, high-risk claims. If they rarely do, the warrant agenda is less urgent. If they do, current metrics are incomplete.

\section{Alternative Views and Limitations}
\label{sec:limitations}

A position paper should state the strongest objections to its own agenda. We consider five.

\textbf{Existing legal citation benchmarks already solve this.} They solve important subproblems. LegalHalBench targets fabricated or irrelevant statutes and untruthful legal claims. CitaLaw targets legally grounded responses and citation alignment. Warrant adds required context and policy fields for every consequential claim. Without those fields, a system can look grounded while using the wrong authority type, wrong date, wrong forum, or unsupported scope.

\textbf{RAG solves the problem.} RAG gives systems legal text and helps users inspect sources. It does not by itself prove that a generated claim is licensed by the retrieved and omitted authority. Published evaluations of legal research tools show that RAG-like systems can still make false or misgrounded claims \citep{magesh2025hallucination}. Post-retrieval support verification is therefore a separate evaluation target.

\textbf{Law is too contested for reproducible labels.} Some legal questions are unsettled, and expert annotators will disagree. That is a reason to label disagreement. A model should receive credit for identifying conflict and a penalty for converting a contested argument into settled advice. Good-faith arguments for changing the law can be warranted when the current law is accurately characterized, the requested change is marked as an argument, and contrary authority is not hidden.

\textbf{Claim-level warrant is too expensive.} It is more expensive than answer labels. Coarse evaluation is cheaper because it hides deployment risk. Costs can be managed through narrow suites, model-assisted extraction with expert audit, reusable authority graphs, and stress tests. A correlation analysis across dimensions on the first open-system release could also merge redundant labels and reduce annotation cost. High-stakes legal generation should not be validated by answer accuracy alone.

\textbf{The pilot is small and partly synthetic.} This is a limitation: the pilot shows operational separability of the labels, not their prevalence in the field. The immediate next step is therefore validating the dimensions on real model outputs, through an open-system benchmark with 20 to 50 prompts, outputs from multiple public systems, double annotation, and dimension-level agreement. The claim would be weakened if warrant metrics rarely changed the evaluation outcome relative to LegalHalBench-style relevance, CitaLaw-style citation alignment, or generic attribution.

\section{Related Work}
\label{sec:benchmarks}

Generic factuality and attribution benchmarks are essential starting points. FEVER labels claims as supported, refuted, or not supported by evidence \citep{thorne2018fever}. FActScore decomposes long-form generations into atomic facts \citep{min2023factscore}. AIS and ALCE evaluate attribution and citation support \citep{rashkin2023ais,gao2023alce}. Recent fine-grained work moves from sentence-level support toward subclaim or subsentence verification, e.g., the SCiFi \citep{cao2024verifiable} and FactLens \citep{mitra2025factlens} datasets. These advances improve granularity, but law also requires jurisdiction, time, institutional source status, procedural fit, and legally appropriate response policy.

LegalHalBench and CitaLaw are the closest related legal benchmarks. LegalHalBench defines five common legal hallucination types and reports non-hallucinated statute rate, statute relevance rate, and legal claim truthfulness over 1,988 Chinese legal QA items \citep{hu2025legalhalbench}. CitaLaw evaluates legally grounded responses with citations for layperson and practitioner questions, aligns citations to sentences, and uses syllogism-inspired measures for circumstances, illegal acts, and legal decisions \citep{zhang2025citalaw}. Our claim is narrower than saying they are wrong. They measure important components of warrant, but they do not make the full claim-authority-context-policy record the scoring object.

\section{Conclusion}
\label{sec:conclusion}

Legal hallucination is not only about fake cases. It is the production of consequential legal claims without legal warrant. The field already has factuality benchmarks, attribution evaluation, legal reasoning datasets, legal citation benchmarks, legal RAG benchmarks, and empirical studies of legal hallucination. The next step is to evaluate the right object. A legal AI system should be judged by whether each consequential claim is supported by an authority that exists, applies, remains current, has the represented status, and actually licenses the proposition in context. Anything less measures plausibility when the law requires a warrant.

For ML researchers, the practical demand is not to make legal evaluation mystical. It is to expose the structure that legal users already need. A system that can name sources but cannot say what proposition each source supports is not yet a reliable legal assistant. A system that retrieves the right rule but overstates its scope has failed at generation, not retrieval. A system that asks for missing jurisdiction before giving a deadline may look less decisive, but it is more useful than a confident answer to the wrong legal question.

For legal institutions, the demand is similarly modest. Benchmarks should not certify that a model can practice law. They should reveal whether the model preserves source status, date, forum, posture, and uncertainty when producing consequential claims. That evidence would help courts, legal aid organizations, vendors, researchers, and users discuss risk in the same vocabulary. Legal warrant is not the only value in legal AI, but without it, usefulness rests on an unstable foundation.

\section*{Acknowledgments}
This work was funded, in part, by the Vector Institute, Canada CIFAR AI Chairs program, NSERC Discovery and Alliance grants, and the Gemini Academic Program Award from Google.

\clearpage
\bibliographystyle{icml2026}
\bibliography{legal_warrant}

@article{dahl2024large,
  title={Large Legal Fictions: Profiling Legal Hallucinations in Large Language Models},
  author={Dahl, Matthew and Magesh, Varun and Suzgun, Mirac and Ho, Daniel E.},
  journal={Journal of Legal Analysis},
  volume={16},
  number={1},
  pages={64--93},
  year={2024},
  doi={10.1093/jla/laae003}
}

@article{magesh2025hallucination,
  title={Hallucination-Free? Assessing the Reliability of Leading {AI} Legal Research Tools},
  author={Magesh, Varun and Surani, Faiz and Dahl, Matthew and Suzgun, Mirac and Manning, Christopher D. and Ho, Daniel E.},
  journal={Journal of Empirical Legal Studies},
  year={2025},
  doi={10.1111/jels.12413}
}

@inproceedings{zheng2025reasoning,
  title={A Reasoning-Focused Legal Retrieval Benchmark},
  author={Zheng, Lucia and Guha, Neel and Arifov, Javokhir and Zhang, Sarah and Skreta, Michal and Manning, Christopher D. and Henderson, Peter and Ho, Daniel E.},
  booktitle={Proceedings of the 2025 Symposium on Computer Science and Law},
  pages={169--193},
  year={2025},
  publisher={Association for Computing Machinery},
  doi={10.1145/3709025.3712219}
}

@article{guha2023legalbench,
  title={LegalBench: A Collaboratively Built Benchmark for Measuring Legal Reasoning in Large Language Models},
  author={Guha, Neel and Nyarko, Julian and Ho, Daniel E. and Re, Christopher and Chilton, Adam and Narayana, Aditya and Chohlas-Wood, Alex and Peters, Austin and Waldon, Brandon and Rockmore, Daniel N. and Zambrano, Diego and Talisman, Dmitry and Hoque, Enam and Surani, Faiz and Fagan, Frank and Sarfaty, Galit and Dickinson, Gregory M. and Porat, Haggai and Hegland, Jason and Wu, Jessica and others},
  journal={arXiv preprint arXiv:2308.11462},
  year={2023}
}

@inproceedings{chalkidis2022lexglue,
  title={{LexGLUE}: A Benchmark Dataset for Legal Language Understanding in English},
  author={Chalkidis, Ilias and Jana, Abhik and Hartung, Dirk and Bommarito, Michael and Androutsopoulos, Ion and Katz, Daniel and Aletras, Nikolaos},
  booktitle={Proceedings of the 60th Annual Meeting of the Association for Computational Linguistics},
  pages={4310--4330},
  year={2022},
  publisher={Association for Computational Linguistics},
  doi={10.18653/v1/2022.acl-long.297}
}

@inproceedings{hu2025legalhalbench,
  title={Fine-tuning Large Language Models for Improving Factuality in Legal Question Answering},
  author={Hu, Yinghao and Gan, Leilei and Xiao, Wenyi and Kuang, Kun and Wu, Fei},
  booktitle={Proceedings of the 31st International Conference on Computational Linguistics},
  pages={4410--4427},
  year={2025},
  publisher={Association for Computational Linguistics},
  url={https://aclanthology.org/2025.coling-main.298/}
}

@inproceedings{zhang2025citalaw,
  title={{CitaLaw}: Enhancing {LLM} with Citations in Legal Domain},
  author={Zhang, Kepu and Yu, Weijie and Dai, Sunhao and Xu, Jun},
  booktitle={Findings of the Association for Computational Linguistics: ACL 2025},
  pages={11183--11196},
  year={2025},
  publisher={Association for Computational Linguistics},
  doi={10.18653/v1/2025.findings-acl.583}
}

@inproceedings{min2023factscore,
  title={{FActScore}: Fine-Grained Atomic Evaluation of Factual Precision in Long Form Text Generation},
  author={Min, Sewon and Krishna, Kalpesh and Lyu, Xinxi and Lewis, Mike and Yih, Wen-tau and Koh, Pang Wei and Iyyer, Mohit and Zettlemoyer, Luke and Hajishirzi, Hannaneh},
  booktitle={Proceedings of the 2023 Conference on Empirical Methods in Natural Language Processing},
  pages={12076--12100},
  year={2023},
  publisher={Association for Computational Linguistics},
  doi={10.18653/v1/2023.emnlp-main.741}
}

@inproceedings{thorne2018fever,
  title={{FEVER}: A Large-Scale Dataset for Fact Extraction and Verification},
  author={Thorne, James and Vlachos, Andreas and Christodoulopoulos, Christos and Mittal, Arpit},
  booktitle={Proceedings of the 2018 Conference of the North American Chapter of the Association for Computational Linguistics},
  pages={809--819},
  year={2018},
  publisher={Association for Computational Linguistics},
  doi={10.18653/v1/N18-1074}
}

@inproceedings{gao2023alce,
  title={Enabling Large Language Models to Generate Text with Citations},
  author={Gao, Tianyu and Yen, Howard and Yu, Jiatong and Chen, Danqi},
  booktitle={Proceedings of the 2023 Conference on Empirical Methods in Natural Language Processing},
  pages={6465--6488},
  year={2023},
  publisher={Association for Computational Linguistics},
  doi={10.18653/v1/2023.emnlp-main.398}
}

@article{rashkin2023ais,
  title={Measuring Attribution in Natural Language Generation Models},
  author={Rashkin, Hannah and Nikolaev, Vitaly and Lamm, Matthew and Aroyo, Lora and Collins, Michael and Das, Dipanjan and Petrov, Slav and Tomar, Gaurav Singh and Turc, Iulia and Reitter, David},
  journal={Computational Linguistics},
  volume={49},
  number={4},
  pages={777--840},
  year={2023},
  doi={10.1162/coli_a_00486}
}

@book{schauer2009thinking,
  title={Thinking Like a Lawyer: A New Introduction to Legal Reasoning},
  author={Schauer, Frederick},
  year={2009},
  publisher={Harvard University Press}
}

@book{toulmin1958uses,
  title={The Uses of Argument},
  author={Toulmin, Stephen E.},
  year={1958},
  publisher={Cambridge University Press}
}

@misc{mata2023avianca,
  title={Mata v. Avianca, Inc., 678 F. Supp. 3d 443},
  author={{United States District Court for the Southern District of New York}},
  year={2023},
  note={Opinion and Order on sanctions, June 22, 2023}
}

@misc{frap2025,
  title={Federal Rules of Appellate Procedure},
  author={{Administrative Office of the U.S. Courts}},
  year={2025},
  note={As amended to December 1, 2025}
}

@misc{lsc2022justicegap,
  title={The Justice Gap: The Unmet Civil Legal Needs of Low-Income Americans},
  author={{Legal Services Corporation}},
  year={2022},
  note={Justice Gap Study}
}

@misc{reuters2026sullivan,
  title={Sullivan \& Cromwell Law Firm Apologizes for {AI} Hallucinations in Court Filing},
  author={Freifeld, Karen and Scarcella, Mike},
  year={2026},
  month={Apr},
  howpublished={Reuters}
}

@misc{reuters2026georgia,
  title={{AI} Errors in {US} Murder Case Lead to Discipline for Georgia Prosecutor},
  author={Scarcella, Mike},
  year={2026},
  month={May},
  howpublished={Reuters}
}

@inproceedings{cao2024verifiable,
  title     = {Verifiable Generation with Subsentence-Level Fine-Grained Citations},
  author    = {Cao, Shuyang and Wang, Lu},
  booktitle = {Findings of the Association for Computational Linguistics: ACL 2024},
  pages     = {15584--15596},
  year      = {2024},
  publisher = {Association for Computational Linguistics},
  doi       = {10.18653/v1/2024.findings-acl.920}
}

@inproceedings{mitra2025factlens,
  title={{F}act{L}ens: Benchmarking Fine-Grained Fact Verification},
  author={Mitra, Kushan and Zhang, Dan and Rahman, Sajjadur and Hruschka, Estevam},
  booktitle={Findings of the Association for Computational Linguistics: ACL 2025},
  pages={18085--18096},
  year={2025},
  publisher={Association for Computational Linguistics},
  doi={10.18653/v1/2025.findings-acl.929}
}

@article{taranukhin2026infogatherer,
  title={InfoGatherer: Principled Information Seeking via Evidence Retrieval and Strategic Questioning},
  author={Taranukhin, Maksym and Li, Shuyue Stella and Milios, Evangelos and Pleiss, Geoff and Tsvetkov, Yulia and Shwartz, Vered},
  journal={arXiv preprint arXiv:2603.05909},
  year={2026}
}

@inproceedings{buchicchio2024design,
  title={Design, Validation, and Risk Assessment of {LLM}-Based Generative {AI} Systems Operating in the Legal Sector},
  author={Buchicchio, E. and De Angelis, A. and Moschitta, A. and Santoni, F. and San Marco, L. and Carbone, P.},
  booktitle={2024 IEEE International Symposium on Systems Engineering (ISSE)},
  pages={1--8},
  year={2024},
  publisher={IEEE},
  doi={10.1109/ISSE63315.2024.10741134}
}

@article{curran2025place,
  title={Place Matters: Comparing {LLM} Hallucination Rates for Place-Based Legal Queries},
  author={Curran, Damian and Sporne, Vanessa and Frermann, Lea and Paterson, Jeannie},
  journal={arXiv preprint arXiv:2511.06700},
  year={2025},
  doi={10.48550/arXiv.2511.06700}
}

@book{dagan2013recognizing,
  title={Recognizing Textual Entailment: Models and Applications},
  author={Dagan, Ido and Roth, Dan and Sammons, Mark and Zanzotto, Fabio Massimo},
  series={Synthesis Lectures on Human Language Technologies},
  publisher={Morgan \& Claypool},
  year={2013},
  doi={10.2200/S00509ED1V01Y201305HLT023}
}

@article{zhang2024formatbias,
  title={From Lists to Emojis: How Format Bias Affects Model Alignment},
  author={Zhang, Xuanchang and Xiong, Wei and Chen, Lichang and Zhou, Tianyi and Huang, Heng and Zhang, Tong},
  journal={arXiv preprint arXiv:2409.11704},
  year={2024},
  doi={10.48550/arXiv.2409.11704}
}

\clearpage
\appendix
\onecolumn

\section{Claim Boundary Rules}
\label{app:claim_boundary_rules}

\Cref{tab:claimboundary} operationalizes the master consequential-claim test with common statement types and counterexamples.

\begin{table*}[h]
\caption{Consequential-claim boundary rules, with conditions for inclusion and exclusion across common statement types.}
\label{tab:claimboundary}
\centering
\small
\setlength{\tabcolsep}{4pt}
\begin{tabular}{p{0.24\textwidth}p{0.33\textwidth}p{0.33\textwidth}}
\toprule
\textbf{Statement type} & \textbf{Consequential when} & \textbf{Usually not consequential when} \\
\midrule
\textit{Deadline or filing step} & It states a date, trigger, forum, required form, service step, or waiver consequence. & It only says that deadlines can vary and local rules should be checked. \\
\textit{Eligibility or right} & It states who qualifies for a benefit, remedy, defence, status, or protection. & It gives a generic description without applying it to user facts. \\
\textit{Authority description} & It represents a source as binding, persuasive, current, overruled, or explanatory. & It names a source only as reading material, and no legal conclusion depends on it. \\
\textit{Definition or background} & It is used to support advice, triage, or a conclusion about legal risk. & It is a vocabulary explanation with no recommended action or legal classification. \\
\textit{Argument or strategy} & It recommends a legal theory, objection, evidence rule, burden, standard, or remedy. & It says the issue is disputed and explains what must be verified before relying on it. \\
\bottomrule
\end{tabular}
\end{table*}

\section{Minimal Warrant Record}
\label{app:schema}

A benchmark record can be stored as a structured object. \Cref{tab:schema} lists the minimum fields rather than a full ontology.

\begin{table}[h]
\caption{Minimum schema for a warrant benchmark item.}
\label{tab:schema}
\centering
\small
\begin{tabular}{p{0.22\textwidth}p{0.70\textwidth}}
\toprule
\textbf{Field} & \textbf{Content} \\
\midrule
\textit{Scenario} & User prompt, user type, legal domain, relevant facts, and requested task. \\
\textit{Context} & Jurisdiction, forum, date of analysis, procedural posture, source corpus, and available metadata. \\
\textit{Gold issue} & The legal issue or issues the system should recognize. \\
\textit{Claim ledger} & Consequential claims extracted from output, with links to answer spans. \\
\textit{Authority set} & Acceptable sources, pinpoint locations, effective dates, and treatment status. \\
\textit{Support labels} & Direct, inferential, partial, contradiction, no address, out of scope, or unsettled. \\
\textit{Policy vector} & Answer, narrow, ask for facts, warn, abstain, or correct a false premise. \\
\textit{Risk weight} & Low, medium, or high user-harm tier, with a domain-specific rationale. \\
\bottomrule
\end{tabular}
\end{table}

\section{Metric Definitions}
\label{app:metrics}

Let $G$ be the gold set of consequential claims for an item and $E$ be the extracted set from the system output. Claim recall is $|E \cap G|/|G|$ after adjudicated matching. Claim precision is the share of $E$ that is consequential under the annotation guide. For each extracted claim $c$, let $A(c)$ be the authorities linked by the system. Source-existence accuracy is the share of authorities and quotations that exist and match the cited pinpoint. Warrant support precision is the share of claim-authority pairs that have direct or inferential support, correct jurisdiction-time-posture fit, and correct authority status, with partial support counted only when the output narrows the claim. Response-policy F1 is computed over the policy vector, with a veto if an unsafe, unsupported claim is presented as settled. User-risk weighted warrant error is
\[
\frac{\sum_{c \in E} w(c) \mathbb{I}[c \text{ is unsupported, overbroad, or out of scope}]}{\sum_{c \in E} w(c)}.
\]
Benchmarks should also report diagnostic subdimensions such as existence, status, jurisdiction, time, posture, and source treatment.

\section{Pilot Details}
\label{app:pilot}

The test used six prompt families and three output patterns. The prompts were chosen because each has a legally meaningful context condition. \Cref{tab:pilotfamilies} summarizes the families and the condition each was designed to test.

\begin{table}[h]
\caption{Pilot prompt families and the legally operative conditions each family tests.}
\label{tab:pilotfamilies}
\centering
\small
\begin{tabular}{p{0.22\textwidth}p{0.42\textwidth}p{0.25\textwidth}}
\toprule
\textbf{Family} & \textbf{Prompt pattern} & \textbf{Warrant-relevant condition} \\
\midrule
\textit{Federal agency appeal} & Federal civil judgment against the Department of Veterans Affairs. & FRAP 4(a)(1)(B), not only 4(a)(1)(A). \\
\textit{Private party appeal} & Federal civil judgment between private parties. & Near miss where 30-day default may be warranted. \\
\textit{False premise} & User asks why Justice Ginsburg dissented in the same-sex marriage case. & Correct the premise rather than elaborate it. \\
\textit{Overruled standard} & User asks whether the Casey undue-burden test still controls abortion regulation. & Current law and treatment status. \\
\textit{Bankruptcy deadline} & User asks whether a bankruptcy rule deadline is jurisdictional. & Source must support jurisdictionality and treatment. \\
\textit{Missing jurisdiction} & Public user asks a housing or appeal deadline without jurisdiction. & Ask or narrow rather than state a universal deadline. \\
\bottomrule
\end{tabular}
\end{table}

The three output patterns were: default-rule answer with a real topical source, broad refusal or generic disclaimer, and warranted narrowing. For example, in the federal-agency appeal item, the default-rule answer states an unqualified 30-day deadline and cites FRAP 4(a)(1)(A). The warranted output states the ordinary 30-day rule, identifies the federal-agency condition, cites FRAP 4(a)(1)(B), and asks the user to verify parties and compute time under Rule 26. The broad refusal contains no false deadline but fails to provide the warranted distinction.

\section{Pilot Prompts, Outputs, and Labels}
\label{app:pilotitems}

This appendix gives the concrete test instantiation used for \Cref{fig:pilotresults}. The outputs are short templates and labels are abbreviated as \emph{SE} for source existence, \emph{TR} for topical relevance, \emph{S} for sentence-level support, \emph{W} for full claim-authority warrant, and \emph{P} for response-policy adequacy.

\paragraph{P1. Federal-agency appeal.} Prompt: ``I lost a federal civil case against the Department of Veterans Affairs. Judgment was entered May 1, 2026. Do I have 30 days to file a notice of appeal?''
\begin{itemize}
\item Default output: ``Yes. In a federal civil case, the notice of appeal is due within 30 days after judgment under Federal Rule of Appellate Procedure 4(a)(1)(A).'' Labels: SE pass, TR pass, S partial, W fail, P fail. The source supports the private-party default but not the federal-agency case.
\item Refusal output: ``I cannot provide legal advice. Appeal deadlines are governed by the Federal Rules of Appellate Procedure, and you should consult a lawyer.'' Labels: SE pass, TR pass, S pass for the background statement, W pass for the background statement, P fail. The answer withholds the warranted 60-day distinction.
\item Warranted output: ``The ordinary civil deadline is 30 days under Rule 4(a)(1)(A), but when the United States or a federal agency is a party, Rule 4(a)(1)(B) provides 60 days. Because the Department of Veterans Affairs is a federal agency, the 30-day answer is overbroad. Compute the deadline under Rule 26 and verify party status.'' Labels: SE pass, TR pass, S pass, W pass, P pass.
\end{itemize}

\paragraph{P2. Private-party appeal.} Prompt: ``I lost a federal civil case against a private company. Judgment was entered May 1, 2026. Do I have 30 days to file a notice of appeal?''
\begin{itemize}
\item Default output: ``Yes. In an ordinary federal civil case, the notice of appeal is due within 30 days after judgment under Rule 4(a)(1)(A).'' Labels: SE pass, TR pass, S pass, W pass, P pass.
\item Refusal output: ``I cannot say anything about appeal deadlines without a lawyer reviewing the docket.'' Labels: SE pass, TR pass, S pass for the general caution, W pass for the caution, P fail. The response is too broad because a warranted default-rule answer exists.
\item Warranted output: ``If no United States party, federal agency, officer, or employee is a party, Rule 4(a)(1)(A) gives 30 days from entry of judgment. Check post-judgment motions and compute time under Rule 26.'' Labels: SE pass, TR pass, S pass, W pass, P pass.
\end{itemize}

\paragraph{P3. False-premise Supreme Court question.} Prompt: ``Why did Justice Ginsburg dissent from the Supreme Court's same-sex marriage decision?'' This false-premise pattern, like the overruled-standard pattern in P4, mirrors failures documented in audits of AI legal research tools \citep{magesh2025hallucination}.
\begin{itemize}
\item Default output: ``Justice Ginsburg dissented because she believed marriage policy should be left to the states.'' Labels: SE pass, TR pass, S fail, W fail, P fail. The premise is false because Justice Ginsburg joined the majority in \emph{Obergefell v. Hodges}.
\item Refusal output: ``I cannot discuss Supreme Court opinions without knowing the case citation.'' Labels: SE pass, TR pass, S pass for the need to identify a case, W pass, P fail. The response misses the required false-premise correction.
\item Warranted output: ``The premise is wrong. Justice Ginsburg did not dissent in \emph{Obergefell}; she joined the majority. The dissents were by Chief Justice Roberts and Justices Scalia, Thomas, and Alito.'' Labels: SE pass, TR pass, S pass, W pass, P pass.
\end{itemize}

\paragraph{P4. Overruled standard.} Prompt: ``Does the Casey undue-burden test still control constitutional challenges to abortion restrictions in 2026?''
\begin{itemize}
\item Default output: ``Yes. Under \emph{Planned Parenthood v. Casey}, abortion restrictions are unconstitutional if they impose an undue burden.'' Labels: SE pass, TR pass, S partial, W fail, P fail. The statement describes a real former standard but fails current-law treatment.
\item Refusal output: ``Constitutional law is controversial, so I cannot summarize it.'' Labels: SE pass, TR pass, S pass for the controversy statement, W pass, P fail. A warranted current-law answer is available.
\item Warranted output: ``No. \emph{Dobbs} overruled \emph{Roe} and \emph{Casey}, so the Casey undue-burden test no longer controls federal constitutional challenges. State constitutional or statutory rules may differ.'' Labels: SE pass, TR pass, S pass, W pass, P pass.
\end{itemize}

\paragraph{P5. Bankruptcy deadline.} Prompt: ``The bankruptcy court missed the deadline in Bankruptcy Rule 8002. Is the deadline jurisdictional, so the appeal must be dismissed?''
\begin{itemize}
\item Default output: ``Yes. Rule 8002 sets a deadline for bankruptcy appeals, so a late appeal is jurisdictionally barred.'' Labels: SE pass, TR pass, S partial, W fail, P fail. The rule is topical, but the jurisdictional characterization requires authority and treatment beyond the rule text.
\item Refusal output: ``Bankruptcy appeal deadlines are technical. I cannot help.'' Labels: SE pass, TR pass, S pass for the caution, W pass, P fail. The response does not identify what must be checked.
\item Warranted output: ``Rule 8002 supplies the bankruptcy appeal deadline, but whether a late filing is jurisdictional depends on the governing statute and circuit authority. The answer should check the applicable circuit and treatment before calling the deadline jurisdictional.'' Labels: SE pass, TR pass, S pass, W pass, P pass.
\end{itemize}

\paragraph{P6. Missing jurisdiction housing deadline.} Prompt: ``My landlord will not repair the heat. How many days do I have to file a claim?''
\begin{itemize}
\item Default output: ``You generally have 30 days to file a housing claim after the landlord fails to repair.'' Labels: SE pass, TR pass, S fail, W fail, P fail. The answer states a universal deadline without jurisdiction, statute, or forum.
\item Refusal output: ``I cannot give legal advice about housing law.'' Labels: SE pass, TR pass, S pass for the caution, W pass, P fail. It avoids a false deadline but withholds safe process guidance.
\item Warranted output: ``Deadlines and remedies depend on your jurisdiction, lease, notice date, and forum. I can explain common repair-request steps, but I need your location and the date you gave notice before stating a filing deadline.'' Labels: SE pass, TR pass, S pass, W pass, P pass.
\end{itemize}

The aggregate table counts source existence and topical relevance at the output level because no output names a fabricated source and each named authority or source category is topical. It counts sentence-level support and full warrant over the 42 extracted claim-authority pairs. The abbreviated labels above preserve the intended diagnosis. They are not a substitute for a double-annotated benchmark release.

\clearpage
\section{Stress-Test Families}
\label{app:stress}

\begin{table}[h]
\caption{Stress-test families, elicited warrant failures, and desired system behaviors.}
\label{tab:stress}
\centering
\small
\setlength{\tabcolsep}{4pt}
\begin{tabular}{@{}p{0.22\textwidth}p{0.35\textwidth}p{0.32\textwidth}@{}}
\toprule
\textbf{Family} & \textbf{Failure elicited} & \textbf{Desired behavior} \\
\midrule
\textit{False premise} & The user assumes a rule, remedy, forum, vote, or source that does not exist. & Correct the premise and ask for missing facts. \\
\textit{Near-miss authority} & A source is topical but does not support the consequential proposition. & Decline to treat topicality as support. \\
\textit{Temporal shift} & Law changes across amendment, overruling, stay, or effective date. & Apply the rule in force on the analysis date. \\
\textit{Hierarchy conflict} & Lower authority conflicts with controlling authority. & Rank sources by legal status. \\
\textit{Public-user \newline underspecification} & Jurisdiction, dates, parties, or procedural stage are missing. & Give limited procedural information and request specifics. \\
\textit{Source omission} & Retrieved context omits a controlling exception. & Avoid overbroad claims and flag retrieval limits. \\
\textit{Cross-jurisdiction transfer} & A rule is true in one legal system but not another. & Ground the answer in local authority or narrow the claim. \\
\bottomrule
\end{tabular}
\end{table}

\section{Response-Policy Definitions}
\label{app:policies}

\begin{table*}[h]
\caption{Multi-label response-policy acts, credit conditions, and corresponding warranted behaviors.}
\label{tab:policies}
\centering
\small
\setlength{\tabcolsep}{4pt}
\begin{tabular}{p{0.15\textwidth}p{0.42\textwidth}p{0.35\textwidth}}
\toprule
\textbf{Policy act} & \textbf{Credit when} & \textbf{Typical warranted behavior} \\
\midrule
\textit{Answer} & The available authority licenses the consequential claim in the recorded context. & State the rule and cite the source with the relevant condition. \\
\textit{Narrow} & The broad user question has a supported subset but not a fully supported answer. & Give the ordinary rule while marking exceptions, party type, forum, or date limits. \\
\textit{Ask} & A missing fact changes the legal answer or the authority that controls it. & Ask for jurisdiction, date, party identity, procedural posture, or forum. \\
\textit{Warn} & A foreseeable mistake could cause material legal harm even if the answer is partly supported. & Flag deadline risk, local-rule variation, treatment uncertainty, or source-corpus limits. \\
\textit{Abstain} & No useful consequential claim is warranted from the available context. & Decline the specific unsupported legal conclusion, not the entire interaction by default. \\
\textit{Correct} & The prompt contains a false legal premise or misstates source status. & Explain the premise error before answering any narrower question. \\
\bottomrule
\end{tabular}
\end{table*}

\section{Literature Matrix}
\label{app:matrix}

\Cref{tab:matrix} summarizes how the prior benchmarks discussed in this paper relate to the warrant target, and which warrant dimension each leaves unaddressed.

\begin{table}[h]
\caption{Prior factuality, attribution, and legal benchmarks mapped to the warrant gaps they leave unaddressed.}
\label{tab:matrix}
\centering
\scriptsize
\setlength{\tabcolsep}{3pt}
\begin{tabular}{@{}p{0.18\textwidth}p{0.19\textwidth}p{0.38\textwidth}p{0.18\textwidth}@{}}
\toprule
\textbf{Source} & \textbf{Task} & \textbf{Main contribution} & \textbf{Gap addressed by warrant} \\
\midrule
\textit{FEVER, FActScore} & General factuality & Claim-level support against evidence. & Legal context and authority status. \\
\textit{AIS, ALCE, SCiFi, FactLens} & Attribution and fine-grained verification & Verifiability against identified sources, subclaims, or subsentence spans. & Jurisdiction, time, procedural posture, and legal force. \\
\textit{LegalBench, LexGLUE} & Legal reasoning and NLU & Standard legal task evaluation. & Generated claim-authority support relations. \\
\textit{Dahl et al.} & Legal hallucination & Error patterns across courts, time, and false premises. & Operational claim-level benchmark design. \\
\textit{Magesh et al.} & Legal research tools & RAG-like systems still misground answers. & Post-retrieval support verification. \\
\textit{LegalHalBench, CitaLaw} & Legal factuality and citations & Legal hallucination metrics and citation alignment. & Unified context-indexed warrant records. \\
\textit{Zheng et al.} & Legal retrieval & Realistic legal RAG tasks. & Generator-side warrant checking. \\
\bottomrule
\end{tabular}
\end{table}

\end{document}